\documentclass[10pt,twocolumn,letterpaper]{article}

\usepackage{wacv}
\usepackage{times}
\usepackage{epsfig}
\usepackage{graphicx}
\usepackage{amsmath}
\usepackage{amssymb}

\usepackage[utf8]{inputenc} 
\usepackage[T1]{fontenc}    
\usepackage{url}            
\usepackage{booktabs}       
\usepackage{amsfonts}       
\usepackage{nicefrac}       
\usepackage{microtype}      

\usepackage{microtype}
\usepackage{wrapfig}
\usepackage{subfigure}
\usepackage{booktabs} 

\usepackage{amsthm}
\usepackage{amsfonts}
\usepackage{bbm}
\usepackage{multirow}
\usepackage{booktabs}
\usepackage{threeparttable}

\usepackage[dvipsnames]{xcolor}
\colorlet{darkgreen}{green!65!black}
\colorlet{darkblue}{blue!75!black}
\colorlet{darkred}{red!80!black}
\definecolor{lightblue}{HTML}{0071bc}
\definecolor{lightgreen}{HTML}{39b54a}

\newcommand{\name} {TGUL}

\newenvironment{Itemize}%
{
\setlength{\leftmargini}{9pt}%
\begin{itemize}%
\setlength{\itemsep}{0pt}%
\setlength{\topsep}{0pt}%
\setlength{\partopsep}{0pt}%
\setlength{\parskip}{0pt}}%
{\end{itemize}}

{
\setlength{\leftmargini}{9pt}%
\begin{enumerate}%
\setlength{\itemsep}{0pt}%
\setlength{\topsep}{0pt}%
\setlength{\partopsep}{0pt}%
\setlength{\parskip}{0pt}}%
{\end{enumerate}}

\graphicspath{{./figures/}}
\newcommand{\deh}[1]{\textcolor{gray}{#1}}
\definecolor{GrayBG}{gray}{0.95}

\def\wacvPaperID{1107} 

\wacvfinalcopy 

\ifwacvfinal
\fi

\ifwacvfinal
\usepackage[breaklinks=true,bookmarks=false]{hyperref}
\else
\usepackage[pagebackref=true,breaklinks=true,colorlinks,bookmarks=false]{hyperref}
\fi

\begin{document}

\title{Unsupervised Learning for Human Sensing Using Radio Signals}

\author{Tianhong Li\thanks{Indicates equal contribution. This work was supported by the GIST-MIT Research Collaboration grant funded by GIST.} \quad Lijie Fan$^{\ast}$ \quad Yuan Yuan$^{\ast}$ \quad Dina Katabi \\ MIT CSAIL}

\maketitle

\ifwacvfinal
\thispagestyle{empty}
\fi

\begin{abstract}
There is a growing literature demonstrating the feasibility of using Radio Frequency (RF) signals to enable key computer vision tasks in the presence of occlusions and poor lighting. It leverages that RF signals traverse walls and occlusions to deliver {\bf through-wall} pose estimation, action recognition, scene captioning, and human re-identification.  However, unlike RGB datasets which can be labeled by human workers, labeling RF signals is a daunting task because such signals are not human interpretable. Yet, it is fairly easy to collect unlabelled RF signals. It would be highly beneficial to use such unlabeled RF data to learn useful representations in an unsupervised manner. Thus, in this paper, we explore the feasibility of adapting RGB-based unsupervised representation learning to RF signals. We show that while contrastive learning has emerged as the main technique for unsupervised representation learning from images and videos, such methods produce poor performance when applied to sensing humans using RF signals.  In contrast, predictive unsupervised learning methods learn high-quality representations  that can be used for multiple downstream RF-based sensing tasks. Our empirical results show that this approach outperforms state-of-the-art RF-based human sensing on various tasks, opening the possibility of unsupervised representation learning from this novel modality. 

\end{abstract}

\vspace{-10pt}
\section{Introduction}
\vspace{-4pt}

RF-based vision has emerged as an attractive research direction that uses Radio Frequency (RF) signals to ``see'' human poses, body shape, and activities through walls and in dark settings \cite{adib2013see,fan2020home,fan2020learning,ogawa2020wi,hsu2019enabling,zhao2018through,zhao2018rf,kotaru2017position,tian2018rf,wang2019person,li2019making,jiang2020towards,ayyalasomayajula2020deep,yang2020rfid,yu2020human,vasisht2018duet,rapczynski2021baseline}.  While visible light can be easily blocked by walls and opaque objects, RF signals in the WiFi range, can traverse such occlusions. RF signals reflect off the human body, provide information to track people, and capture their shape and actions. Previous work has leveraged those properties to detect people and track their walking speed~\cite{adib20143d}. More recent advances in RF-based learning have demonstrated the feasibility of using neural networks to perform various computer vision tasks like pose estimation~\cite{zhao2018rf}, action recognition~\cite{li2019making}, scene captioning~\cite{fan2020home}, and human re-identification~\cite{fan2020learning} using RF signals as the sole input. These systems are capable of working through walls and in dark scenarios, and thus go beyond the limitations faced by RGB-based systems. 

Such RF-based tasks have typically used supervised learning. Yet, unlike RGB datasets, which can be labeled by human workers, labeling RF signals is a daunting task because such signals are not human interpretable. To label RF data, a synchronized human-interpretable stream, like video, must be present to assist the annotator. Specifically, when collecting an RF dataset, researchers deploy an RF device and a camera system, synchronize the data streams from the two systems, and  calibrate their views and positions with respect to each other. The annotator then generates  labels based on the RGB data. However, using RGB data as assistance could only label a small portion of RF data.  In particular, it is hard to use this approach to generate labelled RF datasets of natural living at home since most people would have privacy concerns about deploying cameras in their homes. Also, a single camera has a limited field of view; thus, users typically need to deploy synchronize and calibrate a multi-camera system, introducing significant overhead. Moreover, cameras do not work well in dark settings and with occlusions, which are common indoor scenarios. 

The above limitations motivate the need for learning from unlabelled RF signals.  Unsupervised or self-supervised representation learning has attracted much recent interest and is a rapidly growing research area in computer vision~\cite{he2020momentum,doersch2015unsupervised,ye2019unsupervised,hjelm2018learning,grill2020bootstrap,bachman2019learning,zhuang2019local,misra2020self,han2020memory}.
It refers to learning data representations that capture potential labels of interest and doing so without human supervision. Most unsupervised representation learning methods are designed for RGB data; and no past work has shown the feasibility of unsupervised learning from RF signals.  

\begin{figure}[t]
\centering
\includegraphics[width=\linewidth]{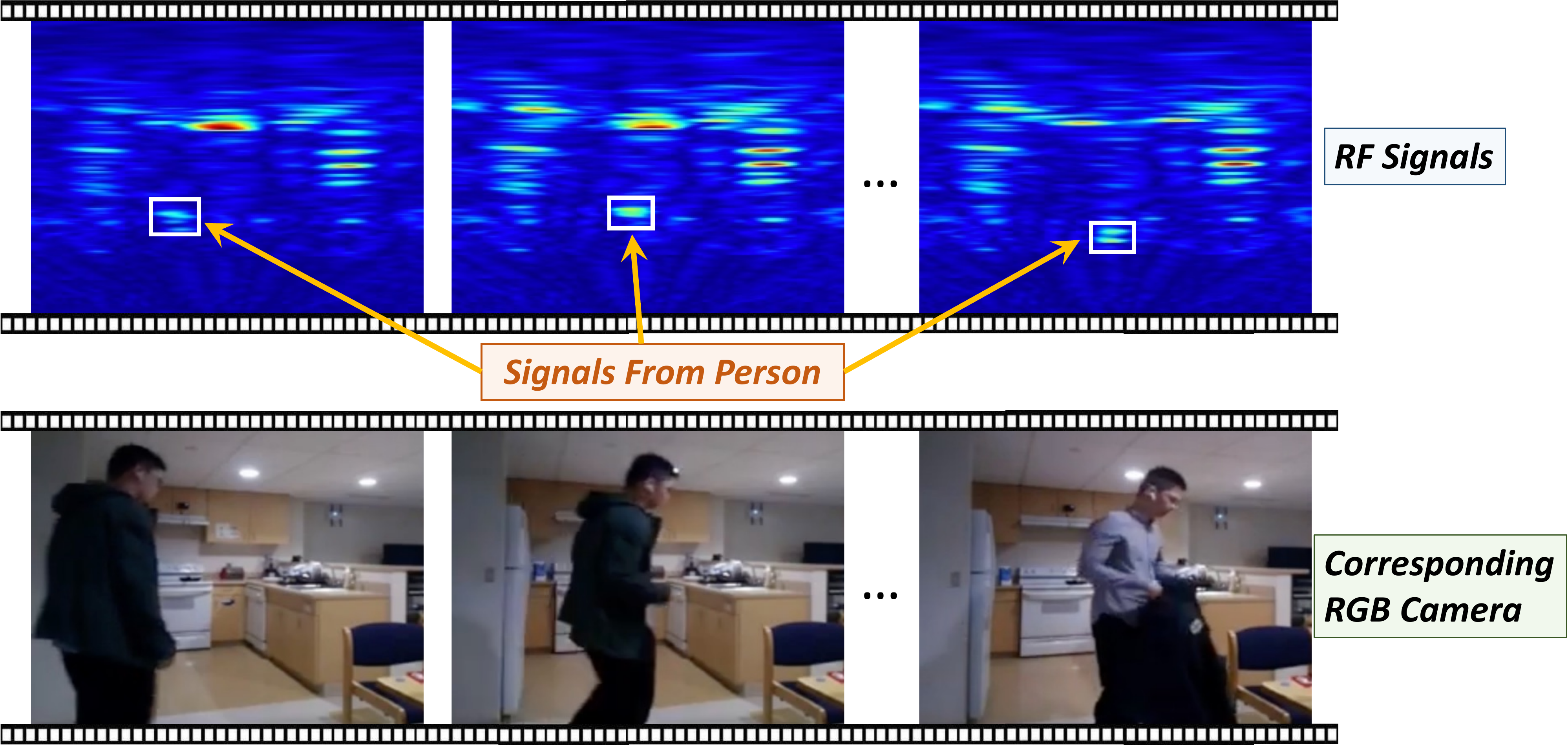}
\caption{\footnotesize{Illustration of RF signals and corresponding RGB images (for illustration \textbf{ONLY}). The signal reflected off the person occupies a small region in the received RF signal. Further, there are many other reflections from other objects.}}\label{fig:teaser}
\vspace{-5pt}
\end{figure}

Applying traditional unsupervised learning methods to RF data is not straight-forward for the following reasons:
\begin{Itemize}
	\item 	Traditional unsupervised methods either rely on strong augmentations, e.g., color jittering in contrastive learning, or require pretext tasks  such as colorizing grey-scale images or predicting image rotation. However, these RGB-specific augmentations and tasks cannot be directly applied to RF signals. RF data is quite different from RGB images. It is obtained by analyzing the signal power reflected from different locations in space. There is no color information in RF signals, and the signal is not invariant to rotation transformation, making most of the existing unsupervised learning methods not directly applicable to RF signals.	
	\item As shown in Figure \ref{fig:teaser}, in RF signals, the information region that corresponds to a person could be extremely small, occupying <1\%. This is because the majority of RF signals are reflected from objects in the environment.  Therefore, directly applying unsupervised learning methods on RF signals could be strongly biased by background noises, and fail to learn useful information about the person. 
	\item RF signals carry much information that is irrelevant to the person or task. Not only human but almost all objects could reflect signals. In particular, only a fraction of an RF signal traverses a wall, and the rest reflects off walls and other objects in the environment. The percentage of the signal that traverses a wall vs. the amount that reflects off it depends on the material and  surface of the wall, and is not easy to estimate. The same can be said about almost all objects in the environment. Furthermore, it is not immediately clear which part of the reflected signal corresponds to the person or people in the scene. This makes it hard for unsupervised learning methods to focus on representations that are useful for human sensing and avoid irrelevant information. 	
\end{Itemize}


In this paper, we introduce a new unsupervised learning framework, Trajectory-Guided Unsupervised Learning (TGUL) to solve the above challenges brought in by RF signals.  We first introduce modified data augmentations and self-supervised tasks that are suitable for RF signals. We then solve the problem of sparse relevant information and eliminate the irrelevant information. To do so, we leverage signal processing techniques that can be applied without any supervision. Specifically we use a radar-based module to detect people and track their trajectories. Then at any point in time, we  zoom in on the region of interest that contains the person, and eliminate signals reflected from other objects. Unsupervised training loss is only added inside this region of interest to avoid the model  learning background noises. 

We adapt TGUL to predictive learning and contrastive learning algorithms. We  evaluate and compare it with state-of-the-art supervised baselines for RF-based sensing tasks, including 3D pose estimation, action recognition, and person re-identification. Our results show that contrastive learning methods could still be strongly biased to learning shortcut information in RF signals, and discarding useful information about the person, while predictive learning methods learn useful information and consistently improve the performance over supervised baselines under the fine-tuning setting. 

To summarize, this paper takes an important step towards extending unsupervised representation learning to new modalities that are hard to interpret by humans. Specifically it makes the following contributions:
\begin{Itemize}
\vspace{-7pt}
    \item The paper introduces trajectory-guided unsupervised learning (\name), a novel unsupervised learning framework for RF signals. We show that \name~is widely applicable to a variety of RF-based human sensing tasks such as pose estimation, action recognition, and person re-identification.
    \item We demonstrate for the first time the possibility of boosting the performance of radio-based human sensing tasks by leveraging unlabeled radio signals. For example, with extra unlabeled RF data, the performance of RF-based pose estimation could be improved by 11.3\%. We further show that even without extra unlabeled data, pre-training the network using \name~can still improve the performance by 5.7\%. 
    \item The paper also shows that state of the art contrastive learning techniques such as SimCLR, MoCo, and BYOL do not work well on radio signals. They tend to learn shortcut information that are contrastive but do not help the downstream tasks. The vulnerability of contrastive learning to shortcuts has been observed in the context of RGB data, where color distribution can be a shortcut that prevents contrastive learning from learning semantic information from images~\cite{chen2020simple}. In the context of RGB data, it is possible to design data augmentations to break such shortcuts, e.g., color jittering.  However,  since RF signals cannot be interpreted by humans, one cannot easily design data augmentations that alter RF signals in a manner that eliminates the shortcut without hampering the original task. As a result, contrastive learning does not perform well on RF signals. 
\vspace{-2pt}
\end{Itemize}

\section{Related Work}

\noindent{\bf RF-based Vision.} 
There is a growing literature that uses radio signals to ``see'' people's poses and activities through walls and in dark settings. Compared with RGB data, radio signals have several advantages: they can traverse walls and occlusions; work in daytime and in darkness; and are not human-interpretable and hence more privacy preserving. Thus, RF-based vision has emerged as an attracting research direction that enables new applications in health care and smart homes~\cite{adib2013see,fan2020learning,ogawa2020wi,hsu2019enabling,zhao2018through,kotaru2017position,tian2018rf,wang2019person,li2019making,jiang2020towards,ayyalasomayajula2020deep,yang2020rfid,yu2020human,vasisht2018duet,rapczynski2021baseline}. 

Previous RF-based vision works include human pose estimation, action recognition, captioning, person re-identifications and so on~\cite{zhao2018through,zhao2018rf,zhao2019through, wang2019person,li2019making,chetty2017low,zhang2018latern,fan2020home,fan2020learning,hsu2019enabling,vasisht2018duet}. However, past works rely on supervised learning. Given the difficulty of annotating RF datasets with ground truth labels, past work is limited to relatively small datasets.

\noindent{\bf Unsupervised Learning from RGB Data.} 
Unsupervised learning methods used in RGB-based vision can be divided into three categories: self-supervision using pretext tasks, predictive approaches, and contrastive approaches. Early work on unsupervised representation learning has focused on designing pretext tasks and training the network to predict their pseudo labels. Such tasks include solving jigsaw puzzles \cite{noroozi2016unsupervised}, colorize grey-scale images \cite{zhang2016colorful} or predicting image rotation \cite{gidaris2018unsupervised}. However, pretext tasks have to be handcrafted and are highly dependent on the properties of RGB data, making them hard to be generalized to other modalities. 

Unsupervised learning based on predictive models including auto-regressive (AR) and auto-encoding (AE) models show less dependence on RGB data. The family of auto-encoders provides a popular framework for unsupervised representation learning using a predictive loss~\cite{hinton2006reducing,pu2016variational,vincent2008extracting}. It trains an encoder to generate low-dimensional latent codes that could reconstruct the entire high-dimensional inputs. There are many types of AEs, such as denoising auto-encoders \cite{vincent2008extracting}, which corrupt the input and let the latent codes reconstruct it, and variational auto-encoders \cite{pu2016variational}, which force the latent codes to follow a prior distribution. However, relevant information in radio signals is typically very sparse. Therefore, traditional AE models can easily ignore the region in RF signal that is relevant to the person and their action. 

Recently, contrastive learning has become widely used for learning effective representations without human supervision. The representations learned with this approach generalize well to downstream tasks, and in some cases can surpass the performance of supervised models  \cite{chen2020simple,chen2020big,chen2020improved,he2020momentum}.  The core idea of contrastive learning is to keep features from positive sample pairs close, and features from negative samples far from each other. Today, multiple successful contrastive learning frameworks exist, with small differences in how they pick negative samples. They include SimCLR \cite{chen2020simple}, the momentum-contrastive approach (MoCo) \cite{he2020momentum}, Contrastive-Multiview-Coding \cite{tian2019contrastive},  and BYOL \cite{grill2020bootstrap}.

However, contrastive learning could suffer from the issue of ``shortcut'' -- i.e., it can learn easy features that are irrelevant to the downstream task, and ignore relevant features~\cite{chen2020simple,tian2020makes}.   For example, in the case of images, the color distribution can be a shortcut that prevents contrastive learning from learning the semantic information in an image~\cite{chen2020simple}. To avoid such shortcut, contrastive learning methods use color jittering as a data augmentation.  In Section \ref{sec:rf-exp}, we show that state-of-the-art contrastive learning methods can fail to learn meaningful representations due to shortcuts in RF signals. However, unlike shortcuts in RGB data, since RF signals cannot be interpreted by humans, one cannot easily design data augmentations that alter the RF signal in a manner that eliminates the shortcut without hampering the original task.

\section{RF Signals Preliminary}
\vspace{-10pt}
\begin{figure}[h]
\centering
\includegraphics[width=.9\linewidth]{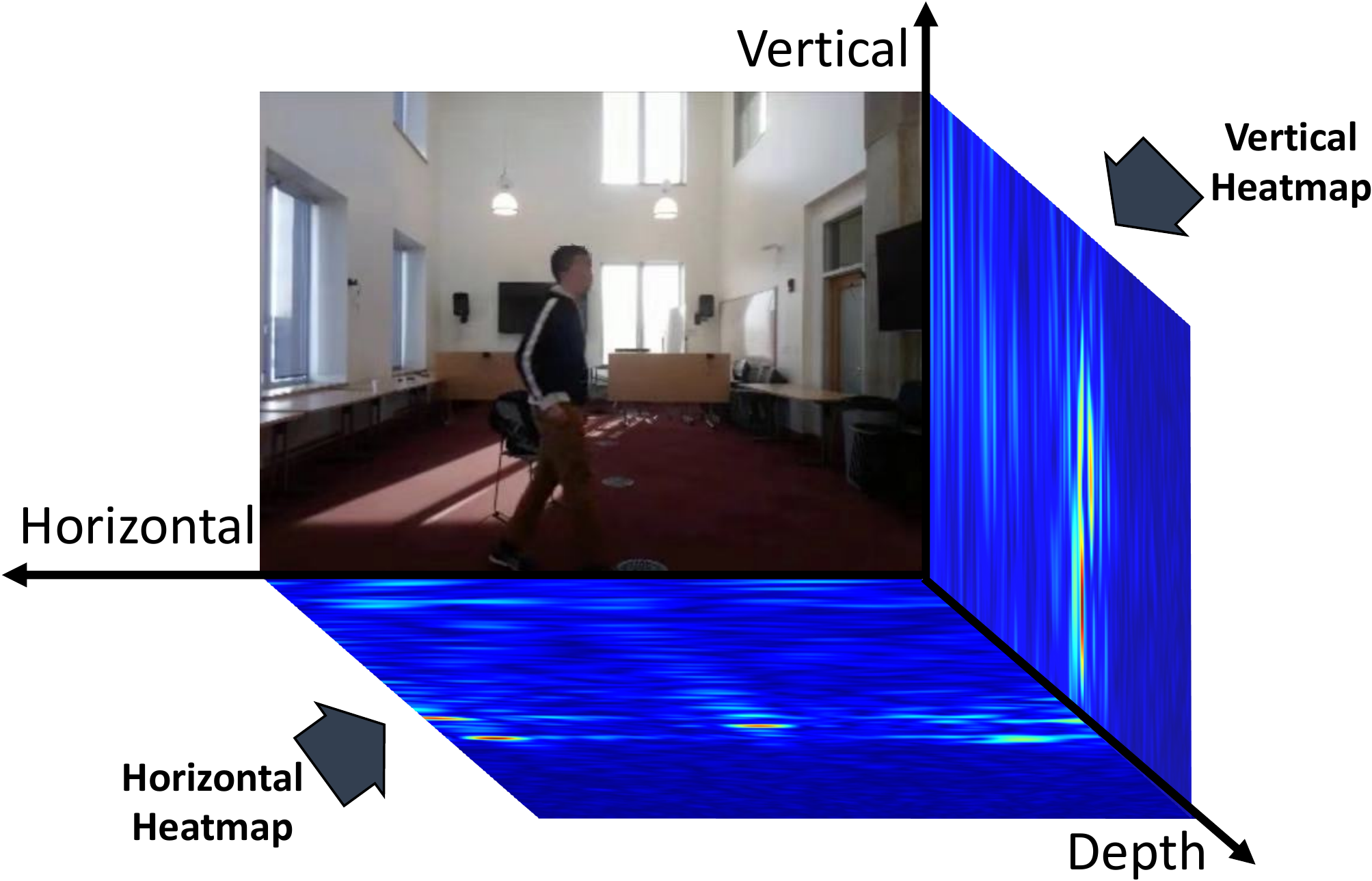}
\caption{\footnotesize{Illustration of RF signals as a pair of horizontal and vertical heatmaps after subtracting static objects, and an RGB image recorded at the same time. Red color refers to high signal power, blue refers to low power.}}\label{fig:rf-heatmaps}
\vspace{-10pt}
\end{figure}

\begin{figure*}[t]
\includegraphics[width=0.95\linewidth]{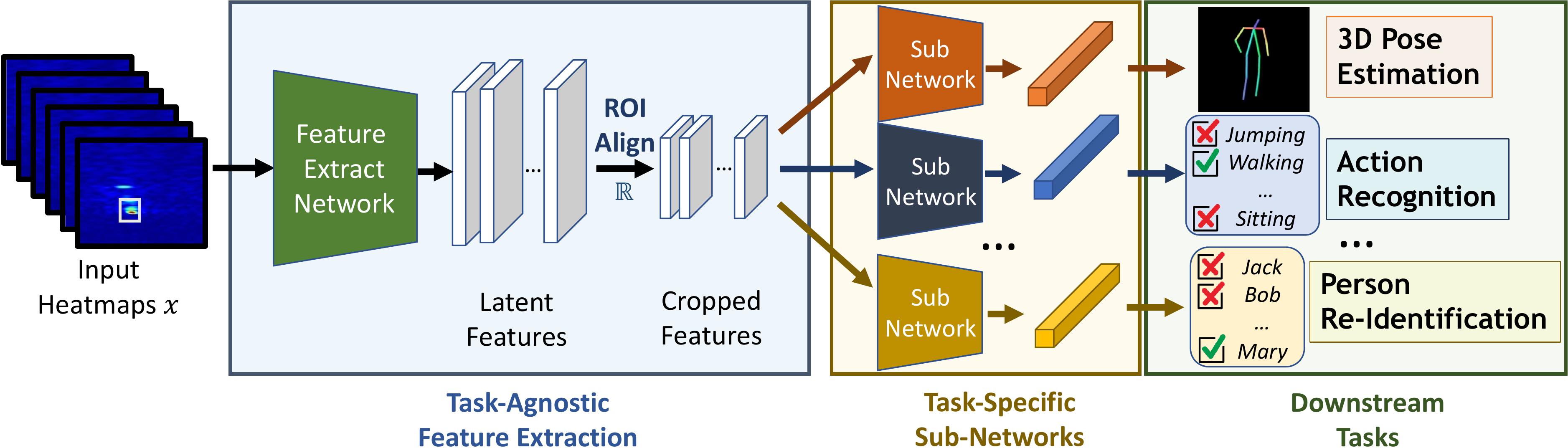}
\centering
\caption{\footnotesize{RF-based human sensing framework. The RF heatmaps are first fed to a Feature Network to extract spatio-temporal features, and ROI Align with the  labeled bounding box is used to crop out the regions with humans. Each downstream task has a unique sub-network to extract task-specific features and predictions.}}\label{fig:rf-framework}
\vspace{-5pt}
\end{figure*}

RF-based human sensing relies on transmitting a low power radio signal and receiving its reflections from the environment and nearby people~\cite{zhao2018through,zhao2018rf,li2019making,chetty2017low,zhang2018latern,fan2020home,zhao2019through,adib2015capturing,vasisht2018duet,tian2018rf,zhao2016emotion,zhao2017learning,hsu2019enabling,fan2020learning}. These technologies typically use a radio device that combines FMCW (Frequency Modulated Continuous Wave) and antenna arrays \cite{adib2015capturing}. This allows the radio to operate like a consumer radar, and separate RF signal reflections from different objects in space. Specifically,  FMCW separates RF reflections based on the distance of the reflecting object, whereas antenna arrays separate reflections based on their spatial direction.

As in past work, we consider a radio device with two antenna arrays: a horizontal one and a vertical one. As a result, the RF signals can be expressed as two heatmaps, as shown in  Fig. \ref{fig:rf-heatmaps}.  The horizontal heatmap is a projection of the radio signals on a plane parallel to the ground, and the vertical heatmap is a projection of the radio signals on a plane perpendicular to the ground.  We use the term RF frame to refer to a pair of horizontal and vertical heatmaps. 




\section{RF-Based Human Sensing Framework}
We choose the following three RF-based sensing tasks to validate the performance of our unsupervised learning method:
\begin{Itemize}
\vspace{-5pt}
\item (a) RF-Based 3D Pose Estimation~\cite{zhao2018through,zhao2018rf}, which uses RF signals to infer the 3D locations of 14 keypoints on the human body: head, neck, shoulders, arms, wrists, hips, knees, and feet.
\item (b) RF-Based action recognition~\cite{li2019making}, which analyzes radio signals to infer human actions and interactions in the presence of occlusions and in poor lighting. 
\item (c) RF-Based Re-Id~\cite{fan2020home}, which recognizes a person-of-interest across different places and times by analyzing the radio signals that bounce off their bodies. 
\vspace{-5pt}
\end{Itemize}

In all these tasks, the neural network has the general structure in Figure~\ref{fig:rf-framework}. The model first uses a spatio-temporal convolutional feature network consisting of 9 residual blocks to extract features from the input RF frames. It then crops out a bounding box around the person in the feature map. Finally, the cropped features are fed to a task-specific sub-network to generate frame-level features. The feature network is the same for all tasks, but different tasks could have different sub-networks specifically designed for that task.

\section{Trajectory Guided Unsupervised Learning}
In this section, we adapt contrastive and predictive unsupervised learning algorithms to learn representations from unlabeled RF signals. 

\begin{figure*}[t]
\begin{center}
\includegraphics[width=0.95\linewidth]{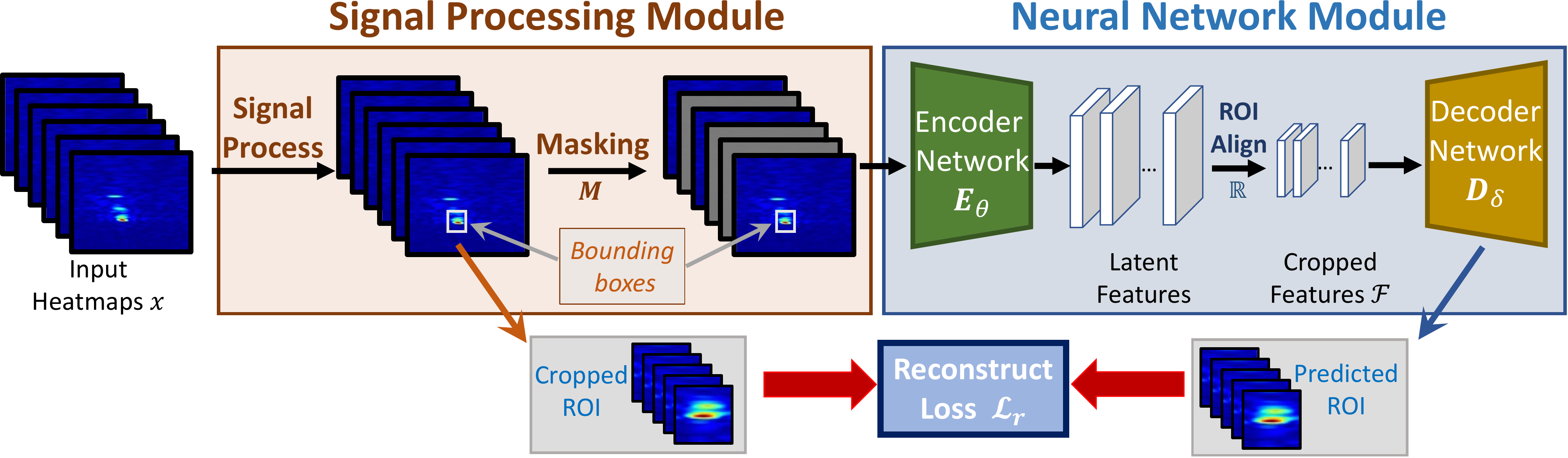}
\end{center}
\caption{\footnotesize{Predictive Learning adapted to RF signals. In the Signal Processing Module (orange box), we use background subtraction to remove static objects and moving detection to obtain bounding boxes for each person in the scene. After this, RF frames are randomly masked and filled with zero. In the Neural Network Module (blue box), a feature extraction network (green box) is applied to extract features from RF inputs, and then an ROI Align module is applied with the  bounding box generated by the signal processing module to crop out regions containing humans. A decoder network is then adopted to predict original input ROIs from cropped features. The reconstruction loss between the cropped ROI of the original heatmap and the predicted ROIs are used to train the model.}}
\label{fig:model-reconstruct}
\vspace{-5pt}
\end{figure*}

\subsection{Signal Processing for ROI Detection}
Unlike standard RGB datasets, where the object of interest typically occupies many of the pixels in an image,  the RF signals reflected from a person only occupy a small portion of an RF heatmap (e.g., < 1\%). Therefore, we need to track and zoom in on the person before we apply unsupervised learning methods to RF signals, otherwise, the unsupervised learning method is likely to learn information about the background instead of the person. 

Traditional approaches for zooming in on individuals or objects in RGB images typically rely on learning a bounding box of this object from ground truth labels,  which are not available in the context of unsupervised learning.  Thus, instead of using methods from computer vision, we leverage the properties of radio signals. Specifically, there are mainly two challenges in detecting the region of interest in the RF signal. The first is to be able to zoom in on the radio signals reflected from a specific location or region in space (presumably the location of the person) and ignore the rest of the RF signal.   To address this issue, we leverage the horizontal and vertical heatmaps, which correspond to the projection of the radio signals on a horizontal plane and a vertical plane. Thus, to zoom in on the signal reflected from a region in space, we can simply crop out the projection of this region on the horizontal and vertical heatmaps.

The second challenge is to determine which region in space contains the person. In RGB data, detecting a person typically requires complex algorithms or neural networks trained using a large dataset. In contrast, we use the fact that FMCW radios can operate as a radar system that tracks and detect moving objects. In most indoor scenarios, people are the only large moving objects. Therefore, we can adapt radar detection algorithms to detect the person and generate their trajectory as they move around. When the person becomes static, the trajectory will stay at the location where the person stops moving. Then it will start tracking the person again when he or she starts moving again. To be more specific, we use a signal-processing technique called WiTrack \cite{adib20143d} to automatically track and generate the trajectory of the person. It first subtracts from heatmap the median over a long period to remove the reflections from static objects in the scene. Then it computes the difference between consecutive frames to detect moving persons and generate their trajectories. We then generate bounding boxes of size 1m$\times$1.5m for the person in each frame based on their trajectory.

\subsection{Predictive Unsupervised Learning from RF}
We first consider learning unsupervised representations by trying to predict how the human body changes the radio signals that bounce of it. 
Our predictive learning framework for RF signals is illustrated in Figure~\ref{fig:model-reconstruct}. Given an input sequence of RF heatmaps $x$ consisting of $T$ RF frames, $x_1, \cdots, x_T$, we first perform the aforementioned signal processing step to detect regions of interest $b_i$ that contain people. We then randomly mask $t$ frames and fill them with zeros to get the masked input $M(x)$. Then the masked input is passed through an encoder network $E$ with parameter $\theta$, an ROI align module $\mathbb{R}$ which crops out regions of interest in feature space using $b_i$, and a decoder network $D$, with parameter $\delta$, to obtain the reconstruction result $ D_{\delta}(\mathbb{R}(E_{\theta}(M(x))_i,b_i))$. The reconstruction loss $\mathcal{L}_{r}$ is defined as the reconstruction error between the cropped ROI from the original input and the reconstructed one $D_{\delta}(\mathbb{R}(E_{\theta}(M(x))_i,b_i))$:
$$
\mathcal{L}_{r} = \sum_{i=1}^T||D_{\delta}(\mathbb{R}(E_{\theta}(M(x))_i,b_i))-\mathbb{R}(x_i, b_i)||_2.
$$

\noindent\textbf{Instantiation.}
For the encoder, we use the same network architecture as prior works \cite{zhao2018rf,fan2020learning,li2019making}, which is a spatio-temporal convolutional feature network consisting of 9 residual blocks. For the decoder, we use a spatial de-convolutional network with 3 residual blocks. For predictive learning, we adapt two standard predictive methods: Auto-encoder \cite{hinton2006reducing} and Inpainting \cite{pathak2016context}. For Auto-encoder, there is no masking operation on the input, i.e., $M$ is an identical mapping. For Inpainting, for each RF sequence consisting of $T=100$ RF frames, we randomly mask out five 5-frame segments within this sequence and fill them with zeros. 


\begin{figure*}[t]
\begin{center}
\includegraphics[width=0.95\linewidth]{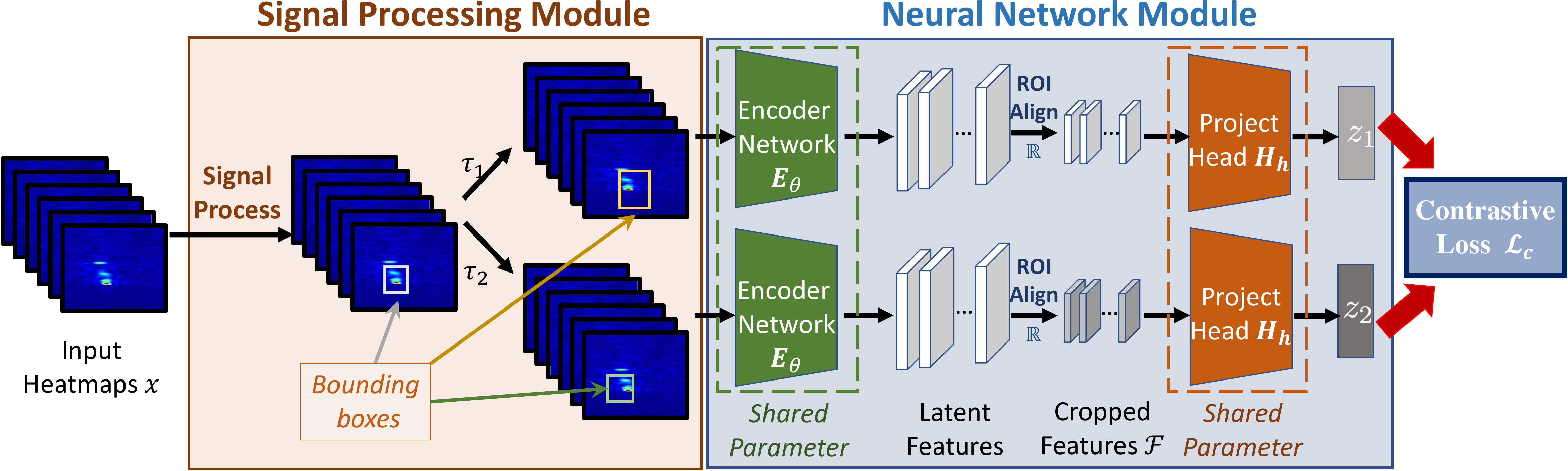}
\end{center}
\caption{\footnotesize{Contrastive Learning on RF signals. In the Signal Processing Module (orange box), we use background subtraction to remove static objects and moving detection to obtain bounding boxes for each person in the scene. After this, different transformation $\tau_1$ and $\tau_2$ to the bounding box are applied to generate positive samples. Both branches share the same feature extraction network (green box), and then use ROI Align with the given bounding box to crop out regions containing humans. A non-linear projection head is used to generate normalized features. Contrastive loss is used to train the model.}}
\label{fig:model-contrast}
\vspace{-10pt}
\end{figure*}

\subsection{Contrastive Learning from RF}
We also consider adapting contrastive learning to RF signals. The contrastive learning framework for RF signals is illustrated in the orange box in Figure~\ref{fig:model-contrast}. For each input RF heatmap sequence $x$ with $T$ frames $x_1, \cdots, x_T$, we first perform the same signal processing step as in predictive learning to detect regions of interest and produce a bounding box $b_i$ for each frame $x_i$. Then we generate a pair of positive samples by performing two data augmentations $\tau_1$ and $\tau_2$ to the generated bounding boxes from the signal processing step, resulting in augmented bounding boxes $\tau_1(b_{i}), \tau_2(b_{i})$. Then we forward the inputs signal with the augmented bounding boxes separately into the encoder $E$ parameterized by $\theta$, an ROI align module $\mathbb{R}$ and a multi-layer nonlinear projection head $H$ parameterized by $h$ to get the latent representations $z_{2i}$ and $z_{2i+1}$ for these two positive samples. Moreover, since human actions are continuous, features within $S=10$ frames of $x_i$ is also taken as positive pairs. We use the commonly used InfoNCE loss \cite{chen2020simple} as the contrastive loss $\mathcal{L}_{c}$. Namely, for a batch of $N$ RF segments containing $NT$ frames $x_{i}^j, i=1,\cdots,T, j=1, \cdots, N $,
\begin{equation*}
\resizebox{0.48\textwidth}{!}{
$
\begin{split}
\mathcal{L}_{c} = -\sum_{j=1}^N\sum_{i=1}^T\sum_{m=1}^S \log \frac{\exp\big(\text{sim}(z^j_{2i}, z^j_{2i-S+2m})/t\big)}{\sum\limits_{l=1}^{N}\sum\limits_{k=1}^{2T} \mathbbm{1}_{k\neq 2i || l\neq j}\exp\big(\text{sim}(z^j_{2i}, z^l_k)/t\big)}
\end{split}
$
}
\end{equation*}

where $\text{sim}(u,v)=u^{T}v / (\|u\|_2 \|v\|_2)$ denotes the dot product between the normalized $u$ and $v$ (i.e., cosine similarity), $t \in \mathbb{R}^{+}$ is a scalar temperature parameter, and $z_{2i},z_{2i+1}$ are the encoded features of positive pairs generated from $x_{i}$, i.e., 
$z_{2i} = H_h(\mathbb{R}(E_{\theta}(x_{i}),\tau_1(b)))$ and 
$z_{2i+1} = H_h(\mathbb{R}(E_{\theta}(x_{i}), \tau_2(b)))$.

Note that the augmentations $\tau_1, \tau_2$ here are crop and resize augmentations on the bounding boxes, instead of the whole input. This is because due to the sparsity of RF signals, performing crop and resize on the whole input may crop out the region of interest.

\noindent\textbf{Instantiation} For the encoder, we use the same network architecture as prior works \cite{zhao2018rf,fan2020learning,li2019making}. For the projection head, we use a spatial convolutional network with one residual block. For negative pairs, we follow the original negative sampling strategy of whatever contrastive learning method \name~build on. Since there is no prior work that applies existing contrastive learning methods to RF data, we implement SimCLR \cite{chen2020simple}, MoCo \cite{he2020momentum}, CPC \cite{henaff2019data} and BYOL \cite{grill2020bootstrap} for RF data by ourselves. The SimCLR implementation is what we just described in Figure \ref{fig:model-contrast}. For MoCo, we use the same data augmentations as SimCLR, except that the positive feature pairs are generated by two feature networks, a normal one and a momentum one. The loss is the same as the one used in \cite{he2020momentum}. For BYOL, we use the same data augmentation operations as SimCLR, and follow the implementation of \cite{grill2020bootstrap} to generate predictions, projections, and loss. For CPC, we follow the design in \cite{henaff2019data}. Specifically, we add a GRU after the RF feature extractor. The output of the GRU at every frame is then used as the context to predict the features in 1.5s in the future using the contrastive loss.

\begin{table*}[t]
\centering
\caption{Evaluation of different models on different RF tasks under fixed feature extractor setting and the relative improvements over network initialized randomly.
$\downarrow$ indicates the smaller the value, the better the performance; $\uparrow$ indicates the larger the value, the better the performance.}
\label{tab:result-fixed}

\begin{threeparttable}
\resizebox{0.95\textwidth}{!}{
\begin{tabular}{l | c |c c|c c c|c c c}
\toprule[1.5pt]
  Tasks & 3D Pose Estimation & \multicolumn{2}{c|}{Action Recognition} & \multicolumn{3}{c|}{Person Re-ID (Campus)} & \multicolumn{3}{c}{Person Re-ID (Home)} \\
\midrule
 \multirow{ 2}{*}{Metrics} &  \multirow{ 2}{*}{Pose \textsc{Err}.$^\downarrow$ (mm)} & \multicolumn{2}{c|}{mAP$^\uparrow$}  & \multirow{ 2}{*}{mAP$^\uparrow$} & \multirow{ 2}{*}{CMC-1$^\uparrow$} & \multirow{ 2}{*}{CMC-5$^\uparrow$} & \multirow{ 2}{*}{mAP$^\uparrow$} & \multirow{ 2}{*}{CMC-1$^\uparrow$} & \multirow{ 2}{*}{CMC-5$^\uparrow$} \\
 & & $\theta=0.1$ & $\theta=0.5$ & & & & & & \\
\midrule\midrule
Random init & 60.1 & 60.5  & 53.3 & 28.1 & 43.8 & 68.8 & 30.1 & 54.2 & 74.6 \\
\midrule
\deh{SimCLR + \name} & \deh{80.5} & \deh{4.2} & \deh{0} & \deh{29.8} & \deh{44.1} & \deh{67.5} & \deh{31.2} & \deh{55.1} & \deh{73.8} \\
\deh{MoCo + \name} & \deh{77.2} & \deh{5.1} & \deh{0.18} & \deh{29.1} & \deh{44.7} & \deh{65.3} & \deh{30.5} & \deh{54.5} & \deh{74.0} \\
\deh{CPC + \name} & \deh{78.7} & \deh{3.6} & \deh{0} & \deh{30.0} & \deh{42.7} & \deh{69.5} & \deh{30.7} & \deh{54.0} & \deh{75.3} \\
\deh{BYOL + \name} & \deh{79.3} & \deh{4.7} & \deh{0} & \deh{29.5} & \deh{44.4} & \deh{66.7} & \deh{30.7} & \deh{54.6} & \deh{73.5} \\
\midrule
Autoencoder & 59.4 & 62.3 & 54.2 & 29.0 & 44.5 & 67.0 & 31.1 & 55.5 & 75.5 \\
Autoencoder + \name & 55.7 & 71.1 & 63.2 & 43.8 & 69.7 & 87.2 & 35.2 & 61.5 & 81.9 \\
Inpainting & 58.0 & 63.9 & 55.4 & 30.2 & 48.1 & 70.5 & 32.8 & 57.7 & 76.5 \\
\textbf{Inpainting + \name} & \textbf{51.1} & \textbf{72.3} & \textbf{65.5} & \textbf{49.8} & \textbf{73.1} & \textbf{90.5} & \textbf{38.5} & \textbf{64.2} & \textbf{84.7} \\
\bf \textsc{Improvement} 
 & \textcolor{darkgreen}{\textbf{+15.0\%}} 
 & \textcolor{darkgreen}{\textbf{+19.5\%}} 
 & \textcolor{darkgreen}{\textbf{+22.9\%}} 
 & \textcolor{darkgreen}{\textbf{+77.2\%}}  
 & \textcolor{darkgreen}{\textbf{+66.9\%}} 
 & \textcolor{darkgreen}{\textbf{+31.5\%}} 
 & \textcolor{darkgreen}{\textbf{+27.9\%}}
 & \textcolor{darkgreen}{\textbf{+18.5\%}} 
 & \textcolor{darkgreen}{\textbf{+13.5\%}}  \\
\bottomrule[1.5pt]
\end{tabular}}
\end{threeparttable}
\vspace{5pt}

\centering
\caption{Evaluation of different models on different RF tasks under fine-tuning setting and the relative improvements over supervised training from random initialized network.
$\downarrow$ indicates the smaller the value, the better the performance; $\uparrow$ indicates the larger the value, the better the performance.}
\label{tab:result-finetune}
\begin{threeparttable}
\resizebox{0.95\textwidth}{!}{
\begin{tabular}{l | c |c c|c c c|c c c}
\toprule[1.5pt]
  Tasks & 3D Pose Estimation & \multicolumn{2}{c|}{Action Recognition} & \multicolumn{3}{c|}{Person Re-ID (Campus)} & \multicolumn{3}{c}{Person Re-ID (Home)} \\
\midrule
 \multirow{ 2}{*}{Metrics} &  \multirow{ 2}{*}{Pose \textsc{Err}.$^\downarrow$ (mm)} & \multicolumn{2}{c|}{mAP$^\uparrow$}  & \multirow{ 2}{*}{mAP$^\uparrow$} & \multirow{ 2}{*}{CMC-1$^\uparrow$} & \multirow{ 2}{*}{CMC-5$^\uparrow$} & \multirow{ 2}{*}{mAP$^\uparrow$} & \multirow{ 2}{*}{CMC-1$^\uparrow$} & \multirow{ 2}{*}{CMC-5$^\uparrow$} \\
 & & $\theta=0.1$ & $\theta=0.5$ & & & & & & \\
\midrule\midrule
Supervised \cite{li2019making,fan2020learning} & 38.4 & 90.1  & 87.8 & 59.5 & 82.1 & 95.5 & 46.4 & 74.6 & 89.5 \\
\midrule
\deh{SimCLR + \name} & \deh{38.8} & \deh{89.8} & \deh{87.4} & \deh{59.0} & \deh{81.7} & \deh{94.1} & \deh{45.9} & \deh{73.8} & \deh{88.5} \\
\deh{MoCo + \name} & \deh{38.3} & \deh{89.7} & \deh{87.2} & \deh{59.3} & \deh{82.0} & \deh{94.5} & \deh{46.4} & \deh{74.3} & \deh{89.7} \\
\deh{CPC + \name} & \deh{38.6} & \deh{89.9} & \deh{87.5} & \deh{59.4} & \deh{81.5} & \deh{94.0} & \deh{46.0} & \deh{74.5} & \deh{89.1} \\
\deh{BYOL + \name} & \deh{38.5} & \deh{89.7} & \deh{87.2} & \deh{59.4} & \deh{81.9} & \deh{94.5} & \deh{46.6} & \deh{74.5} & \deh{89.5} \\
\midrule
Autoencoder & 38.5 &  90.0 & 87.7 & 59.1 & 81.9 & 95.5 & 45.9 & 74.2 & 88.6\\
Autoencoder + \name & 37.5 &  91.2 & 87.9 & 59.7 & 82.8 & 95.5 & 46.8 & 74.6 & 89.8\\
Inpainting & 38.2 &  90.5 & 88.0 & 59.3 & 82.1 & 95.7 & 46.2 & 74.4 & 89.2\\
\textbf{Inpainting + \name} & \textbf{36.2} &  \textbf{91.7} & \textbf{88.7} & \textbf{60.1} & \textbf{83.3} & \textbf{95.9} & \textbf{47.5} & \textbf{75.3} & \textbf{90.3}\\
\bf \textsc{Improvement} 
 & \textcolor{darkgreen}{\textbf{+5.7\%}} 
 & \textcolor{darkgreen}{\textbf{+1.8\%}} 
 & \textcolor{darkgreen}{\textbf{+1.0\%}}  
 & \textcolor{darkgreen}{\textbf{+1.0\%}}  
 & \textcolor{darkgreen}{\textbf{+1.5\%}}  
 & \textcolor{darkgreen}{\textbf{+0.4\%}}  
 & \textcolor{darkgreen}{\textbf{+2.4\%}}  
 & \textcolor{darkgreen}{\textbf{+0.9\%}}  
 & \textcolor{darkgreen}{\textbf{+0.9\%}}  \\

\bottomrule[1.5pt]
\end{tabular}}
\end{threeparttable}
\vspace{-15pt}
\end{table*}

\section{Experiments}
\label{sec:rf-exp}

\noindent\textbf{RF Datasets.} We use two real-world RF datasets used in past work: RF-MMD \cite{li2019making} and RRD \cite{fan2020learning}. RF-MMD is an RF-based 3D pose-estimation and action-recognition dataset consisting of 25 hours of RF data with 30 volunteers performing 35 actions in 10 different indoor environments. RRD is an RF-based person re-identification dataset. It contains  863 RF sequences from 100 different identities at 5 different locations on a campus, and 6305 RF sequences from 38 different identities in 19 homes.

\noindent\textbf{Evaluation Metrics.}  We use the performance metrics used by past work for each task. Specifically, for 3D pose estimation, we use the average $l_2$ distance between 14 predicted keypoints and their ground-truth locations as the evaluation metric. For action recognition, we use mean average precision (mAP) at different intersection-over-union (IoU) thresholds $\theta$ to evaluate the accuracy of the model in detecting and classifying an action event. For person re-identification, we separate the dataset into query set and gallery set. The query samples and gallery samples are then encoded to feature vectors. We calculate the cosine distance between the features of each query sample and each gallery sample, and rank the distance to get the top-N closest gallery samples for each query sample. Based on the ranking results, we compute the mean average precision (mAP) and the cumulative matching characteristic (CMC) at rank-1 and rank-5.

\noindent\textbf{Evaluation Settings.} We evaluate the performance under both fixed feature extractor setting and fine-tuning setting. In the fixed feature extractor setting, we use the pre-trained RF feature extractor as initialization of the RF feature extractor for each downstream task, and only those task-specific sub-networks (Fig. \ref{fig:rf-framework}) are trained during downstream task training. In the fine-tuning setting, we fine-tune the whole model on the downstream task. 

\noindent\textbf{Training Details.} For all approaches, we train the network for 50 epochs with a batch size of 128, using the Adam optimizer \cite{kingma2014adam} with 1e-3 learning rate and 1e-5 weight decay. We use PyTorch \cite{paszke2019pytorch} for implementations.

\subsection{Main Results}

We evaluate our unsupervised learning method on multiple downstream RF-based tasks, and report the results in Tables \ref{tab:result-fixed} and \ref{tab:result-finetune}. Note that contrastive leaning methods could not be directly applied to RF signals without \name~because the RGB-based augmentations is not directly applicable to RF signals. Table \ref{tab:result-fixed} shows the results of the fixed feature extractor setting. The table reveals two findings.  First, predictive pre-training using TGUL learns useful unsupervised representations and delivers 15.0\% relative improvement in 3D pose estimation error, 19.5\% relative improvement in action recognition mAP at $\theta=0.1$, 77.2\% relative improvement in person ReID mAP on RRD-Campus, and 27.9\% relative improvement on RRD-Home. Further, inpainting, i.e., predicting missing RF signals, is more powerful than autoencoding though both deliver gains. 

Second, the table shows that  contrastive pre-training performs worse on all the downstream tasks than predictive pre-training, and even worse than random initialization for the 3D pose estimation and action recognition tasks. This is because contrastive learning learns shortcut semantics irrelevant to the tasks of interest. We further investigate this issue of shortcut in Section \ref{sec:feat-viz}. 

Table \ref{tab:result-finetune} shows the results of the fine-tuning setting. The table shows that initializing supervised models with Inpainting+TGUL's representations can achieve 5.7\% relative improvement in 3D pose estimation error, 1.8\% relative improvement in action recognition mAP at $\theta=0.1$, 1.0\% relative improvement in person ReID mAP on RRD-Campus, and 2.4\% relative improvement on RRD-Home. In contrast,  initializing supervised models with contrastive representations is not useful. 

Tables \ref{tab:result-fixed} and \ref{tab:result-finetune} also demonstrate the effectiveness of \name. Directly applying Autoencoder or Inpainting on RF signals only brings limited improvements over randomly initialized network under fixed feature network setting, and cannot improve the performance under fine-tuning setting. On the other hand, Autoencoder and Inpainting with TGUL achieves much better performance under both settings. This is because \name~is effective in forcing the network to focus on the person and pay less attention to other regions which contain mostly noises, and thus achieves better performance than without this guidance.

Overall, the results show that our framework is the first unsupervised method to boost the performance of RF-based human sensing, bringing the gains of unsupervised learning to this new data modality.



\subsection{Feature Visualization}
\label{sec:feat-viz}

\begin{figure}[h]
\vspace{-5pt}
\begin{center}
\begin{tabular}{ccc}
\includegraphics[width=0.148\textwidth]{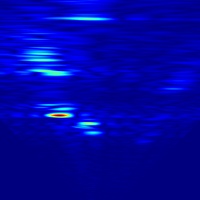} & \hspace*{-0.12in}
\label{fig:intro_rf}
\includegraphics[width=0.148\textwidth]{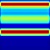} & \hspace*{-0.12in}
\includegraphics[width=0.148\textwidth]{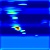}\\
{\footnotesize (a) Original Heatmap} & \hspace*{-0.12in} {\footnotesize (b) CL Features} & \hspace*{-0.12in} {\footnotesize(c) PL features}
\end{tabular}
\end{center}
\vspace{-5pt}
\caption{Contrastive learning (CL) can be biased to shortcut and discard useful information in the input, while predictive learning (PL) can preserve the information in the input.}
\label{fig:feat-viz}
\vspace{-15pt}
\end{figure}

As shown in Table \ref{tab:result-fixed} and \ref{tab:result-finetune}, contrastive pre-training methods perform poorly and even worse than random initialization of the feature network. This is because contrastive learning is likely to exploit shortcut information (e.g., the distance of the person to the RF device) and discard other useful information for downstream tasks. In Figure \ref{fig:feat-viz}, we visualize the features generated by feature networks pre-trained using contrastive learning and predictive learning. As shown in the figure, the features network pre-trained using contrastive learning discard most of the information in the input. On the other hand, the feature network pre-trained using predictive learning can preserve and abstract the information about the person in the input and thus improve the performance.

\subsection{Ablation Studies}
We perform ablation studies on our proposed unsupervised learning framework. All ablation studies are performed on the RF-MMD \cite{li2019making} dataset with the downstream task of pose estimation, where we report the average error across all keypoints.

\textbf{Masking Strategy}: When applied to RF signals, the Inpainting task is performed by masking some frames in a short window of RF signals and reconstructing them from the remaining frames. We compare the performance of Inpainting with different masking strategies. The input RF signal is a 100-frames RF sequence. We compare 3 different masking strategies: center segment (25 frames in the center of the input), random segments (five 5-frame segments at random positions of input), and random (each frame is masked with 0.25 probability). As shown in the results below, the random segments masking strategy performs the best. 

\begin{table}[h]
\centering
\caption{Pose error (the lower the better) with different masking strategies.}
\label{tab:result-pose}
\begin{tabular}{c |@{\hspace{0.2cm}} c c @{\hspace{0.2cm}} c @{\hspace{0.2cm}}c @{\hspace{0.2cm}}c}\toprule[1.5pt]
 Mask Strategy & Pose \textsc{Err.}$^\downarrow$ (mm) \\
\midrule
Center Segment & 36.8 \\
Random Segments & \textbf{36.2} \\
Random  & 36.4\\
\bottomrule[1.5pt]
\end{tabular}
\end{table}

\textbf{Size of Masked Segment}: Next, we study the influence of the size of each masked segment. Note that when the mask size equals 0, no frame is masked, and the inpainting task becomes a normal auto-encoder. In all of these experiments, we mask 5 segments and follow the strategy of randomly locating the masked segments in the input. As shown in Table \ref{tab:result-prob}, the best performance is for masking 5 consecutive frames, i.e., a segment of size 5. 

\begin{table}[h]
\centering
\caption{Pose error with different sizes of masked segments.}
\label{tab:result-prob}
\begin{tabular}{c |@{\hspace{0.2cm}} c c @{\hspace{0.2cm}} c @{\hspace{0.2cm}}c @{\hspace{0.2cm}}c}\toprule[1.5pt]
 Masking Size & 0 & 1 & 3 & 5 & 10 \\
\midrule
Pose \textsc{Err.}$^\downarrow$ (mm) & 37.5 & 37.0 & 36.3 & \textbf{36.2} & 36.7 \\

\bottomrule[1.5pt]
\end{tabular}
\end{table}

\textbf{Benefits of Unlabeled Data for RF-Based Sensing}: By enabling RF-based tasks to leverage unsupervised representation learning, our framework allows RF-based sensing to benefit from a large amount of unlabeled RF data. To evaluate the benefit of such unlabeled data,  we simulate the scenario where only a handful of labeled RF data is available and a large amount of RF data is unlabeled data. Specifically, we randomly select 10\% of the training set of RF-MMD to be RF-MMD-S to serves as the small labeled dataset. We compare the performance of Inpainting when it is pre-trained on RF-MMD-S and RF-MMD and finetuned on RF-MMD-S. 

\begin{table}[h]
\centering
\caption{Performance of Inpainting on RF-MMD with a small amount of labeled data. The results in the table use RF-MMD-S for finetuning, which is a randomly selected 10\% subset of RF-MMD which is used for unsupervised pre-training. The rest of the dataset is used without labels. The table shows that our framework can further improve the performance with more unlabeled data.}
\label{tab:result-data}
\begin{tabular}{c |@{\hspace{0.2cm}} c c @{\hspace{0.2cm}} c @{\hspace{0.2cm}}c @{\hspace{0.2cm}}c}\toprule[1.5pt]
Methods & Pose \textsc{Err.}$^\downarrow$ (mm) \\
\midrule
Training from scratch (RF-MMD-S) & 48.7 \\
Inpainting on RF-MMD-S+finetune & 46.1\\
Inpainting on RF-MMD+finetune & 43.2\\
\bottomrule[1.5pt]
\end{tabular}

\end{table}

As shown in Table \ref{tab:result-data}, our framework can improve the performance of RF-based 3D pose estimation by 5.3\% without using any additional unlabeled data. This is because our framework can learn a general representation of RF signals and thus provide better generalization ability. Indeed, our framework improves the pose error further by another 6.0\% by leveraging the unlabeled data. This demonstrates the potential of using our framework to leverage large-scale unlabeled RF data to improve the performance of RF-based human sensing methods.

\vspace{-5pt}
\section{Conclusion}
\vspace{-3pt}
We introduced \name, the first unsupervised learning framework for RF-based human sensing tasks. We adapted state-of-the-art unsupervised learning methods from RGB data to RF data by leveraging radar technology to focus unsupervised learning on the person's trajectory, and design data augmentations suitable for RF signals. We showed that contrastive learning could be strongly biased by shortcuts in RF data, making it a poor choice for this modality. We further demonstrate the potential of unsupervised learning methods based on predictive tasks for learning useful representations from RF data. Extensive empirical results on multiple RF datasets and tasks show that \name~could consistently improve the performance of RF-based sensing models over supervised learning baselines. 

{\small
\bibliographystyle{ieee_fullname}
\bibliography{main}

\begin{thebibliography}{10}\itemsep=-1pt

\bibitem{adib2015capturing}
Fadel Adib, Chen-Yu Hsu, Hongzi Mao, Dina Katabi, and Fr{\'e}do Durand.
\newblock Capturing the human figure through a wall.
\newblock {\em ACM Transactions on Graphics (TOG)}, 34(6):1--13, 2015.

\bibitem{adib20143d}
Fadel Adib, Zach Kabelac, Dina Katabi, and Robert~C Miller.
\newblock 3d tracking via body radio reflections.
\newblock In {\em 11th $\{$USENIX$\}$ Symposium on Networked Systems Design and
  Implementation ($\{$NSDI$\}$ 14)}, pages 317--329, 2014.

\bibitem{adib2013see}
Fadel Adib and Dina Katabi.
\newblock See through walls with wifi!
\newblock In {\em Proceedings of the ACM SIGCOMM 2013 conference on SIGCOMM},
  pages 75--86, 2013.

\bibitem{ayyalasomayajula2020deep}
Roshan Ayyalasomayajula, Aditya Arun, Chenfeng Wu, Sanatan Sharma,
  Abhishek~Rajkumar Sethi, Deepak Vasisht, and Dinesh Bharadia.
\newblock Deep learning based wireless localization for indoor navigation.
\newblock In {\em Proceedings of the 26th Annual International Conference on
  Mobile Computing and Networking}, pages 1--14, 2020.

\bibitem{bachman2019learning}
Philip Bachman, R~Devon Hjelm, and William Buchwalter.
\newblock Learning representations by maximizing mutual information across
  views.
\newblock {\em arXiv preprint arXiv:1906.00910}, 2019.

\bibitem{chen2020simple}
Ting Chen, Simon Kornblith, Mohammad Norouzi, and Geoffrey Hinton.
\newblock A simple framework for contrastive learning of visual
  representations.
\newblock {\em arXiv preprint arXiv:2002.05709}, 2020.

\bibitem{chen2020big}
Ting Chen, Simon Kornblith, Kevin Swersky, Mohammad Norouzi, and Geoffrey~E
  Hinton.
\newblock Big self-supervised models are strong semi-supervised learners.
\newblock {\em Advances in Neural Information Processing Systems}, 33, 2020.

\bibitem{chen2020improved}
Xinlei Chen, Haoqi Fan, Ross Girshick, and Kaiming He.
\newblock Improved baselines with momentum contrastive learning.
\newblock {\em arXiv preprint arXiv:2003.04297}, 2020.

\bibitem{chetty2017low}
Kevin Chetty, Qingchao Chen, Matthew Ritchie, and Karl Woodbridge.
\newblock A low-cost through-the-wall fmcw radar for stand-off operation and
  activity detection.
\newblock In {\em Radar Sensor Technology XXI}, volume 10188, page 1018808.
  International Society for Optics and Photonics, 2017.

\bibitem{doersch2015unsupervised}
Carl Doersch, Abhinav Gupta, and Alexei~A Efros.
\newblock Unsupervised visual representation learning by context prediction.
\newblock In {\em Proceedings of the IEEE international conference on computer
  vision}, pages 1422--1430, 2015.

\bibitem{fan2020learning}
Lijie Fan, Tianhong Li, Rongyao Fang, Rumen Hristov, Yuan Yuan, and Dina
  Katabi.
\newblock Learning longterm representations for person re-identification using
  radio signals.
\newblock In {\em Proceedings of the IEEE/CVF Conference on Computer Vision and
  Pattern Recognition}, pages 10699--10709, 2020.

\bibitem{fan2020home}
Lijie Fan, Tianhong Li, Yuan Yuan, and Dina Katabi.
\newblock In-home daily-life captioning using radio signals.
\newblock {\em arXiv preprint arXiv:2008.10966}, 2020.

\bibitem{gidaris2018unsupervised}
Spyros Gidaris, Praveer Singh, and Nikos Komodakis.
\newblock Unsupervised representation learning by predicting image rotations.
\newblock {\em arXiv preprint arXiv:1803.07728}, 2018.

\bibitem{grill2020bootstrap}
Jean-Bastien Grill, Florian Strub, Florent Altch{\'e}, Corentin Tallec,
  Pierre~H Richemond, Elena Buchatskaya, Carl Doersch, Bernardo~Avila Pires,
  Zhaohan~Daniel Guo, Mohammad~Gheshlaghi Azar, et~al.
\newblock Bootstrap your own latent: A new approach to self-supervised
  learning.
\newblock {\em arXiv preprint arXiv:2006.07733}, 2020.

\bibitem{han2020memory}
Tengda Han, Weidi Xie, and Andrew Zisserman.
\newblock Memory-augmented dense predictive coding for video representation
  learning.
\newblock {\em arXiv preprint arXiv:2008.01065}, 2020.

\bibitem{he2020momentum}
Kaiming He, Haoqi Fan, Yuxin Wu, Saining Xie, and Ross Girshick.
\newblock Momentum contrast for unsupervised visual representation learning.
\newblock In {\em Proceedings of the IEEE/CVF Conference on Computer Vision and
  Pattern Recognition}, pages 9729--9738, 2020.

\bibitem{henaff2019data}
Olivier~J H{\'e}naff, Aravind Srinivas, Jeffrey De~Fauw, Ali Razavi, Carl
  Doersch, SM Eslami, and Aaron van~den Oord.
\newblock Data-efficient image recognition with contrastive predictive coding.
\newblock {\em arXiv preprint arXiv:1905.09272}, 2019.

\bibitem{hinton2006reducing}
Geoffrey~E Hinton and Ruslan~R Salakhutdinov.
\newblock Reducing the dimensionality of data with neural networks.
\newblock {\em science}, 313(5786):504--507, 2006.

\bibitem{hjelm2018learning}
R~Devon Hjelm, Alex Fedorov, Samuel Lavoie-Marchildon, Karan Grewal, Phil
  Bachman, Adam Trischler, and Yoshua Bengio.
\newblock Learning deep representations by mutual information estimation and
  maximization.
\newblock {\em arXiv preprint arXiv:1808.06670}, 2018.

\bibitem{hsu2019enabling}
Chen-Yu Hsu, Rumen Hristov, Guang-He Lee, Mingmin Zhao, and Dina Katabi.
\newblock Enabling identification and behavioral sensing in homes using radio
  reflections.
\newblock In {\em Proceedings of the 2019 CHI Conference on Human Factors in
  Computing Systems}, pages 1--13, 2019.

\bibitem{jiang2020towards}
Wenjun Jiang, Hongfei Xue, Chenglin Miao, Shiyang Wang, Sen Lin, Chong Tian,
  Srinivasan Murali, Haochen Hu, Zhi Sun, and Lu Su.
\newblock Towards 3d human pose construction using wifi.
\newblock In {\em Proceedings of the 26th Annual International Conference on
  Mobile Computing and Networking}, pages 1--14, 2020.

\bibitem{kingma2014adam}
Diederik~P Kingma and Jimmy Ba.
\newblock Adam: A method for stochastic optimization.
\newblock {\em arXiv preprint arXiv:1412.6980}, 2014.

\bibitem{kotaru2017position}
Manikanta Kotaru and Sachin Katti.
\newblock Position tracking for virtual reality using commodity wifi.
\newblock In {\em Proceedings of the IEEE Conference on Computer Vision and
  Pattern Recognition}, pages 68--78, 2017.

\bibitem{li2019making}
Tianhong Li, Lijie Fan, Mingmin Zhao, Yingcheng Liu, and Dina Katabi.
\newblock Making the invisible visible: Action recognition through walls and
  occlusions.
\newblock In {\em Proceedings of the IEEE International Conference on Computer
  Vision}, pages 872--881, 2019.

\bibitem{misra2020self}
Ishan Misra and Laurens van~der Maaten.
\newblock Self-supervised learning of pretext-invariant representations.
\newblock In {\em Proceedings of the IEEE/CVF Conference on Computer Vision and
  Pattern Recognition}, pages 6707--6717, 2020.

\bibitem{noroozi2016unsupervised}
Mehdi Noroozi and Paolo Favaro.
\newblock Unsupervised learning of visual representations by solving jigsaw
  puzzles.
\newblock In {\em European Conference on Computer Vision}, pages 69--84.
  Springer, 2016.

\bibitem{ogawa2020wi}
Masakatsu Ogawa and Hirofumi Munetomo.
\newblock Wi-fi csi-based outdoor human flow prediction using a support vector
  machine.
\newblock {\em Sensors}, 20(7):2141, 2020.

\bibitem{paszke2019pytorch}
Adam Paszke, Sam Gross, Francisco Massa, Adam Lerer, James Bradbury, Gregory
  Chanan, Trevor Killeen, Zeming Lin, Natalia Gimelshein, Luca Antiga, et~al.
\newblock Pytorch: An imperative style, high-performance deep learning library.
\newblock {\em arXiv preprint arXiv:1912.01703}, 2019.

\bibitem{pathak2016context}
Deepak Pathak, Philipp Krahenbuhl, Jeff Donahue, Trevor Darrell, and Alexei~A
  Efros.
\newblock Context encoders: Feature learning by inpainting.
\newblock In {\em Proceedings of the IEEE conference on computer vision and
  pattern recognition}, pages 2536--2544, 2016.

\bibitem{pu2016variational}
Yunchen Pu, Zhe Gan, Ricardo Henao, Xin Yuan, Chunyuan Li, Andrew Stevens, and
  Lawrence Carin.
\newblock Variational autoencoder for deep learning of images, labels and
  captions.
\newblock {\em arXiv preprint arXiv:1609.08976}, 2016.

\bibitem{rapczynski2021baseline}
Micha{\l} Rapczy{\'n}ski, Philipp Werner, Sebastian Handrich, and Ayoub
  Al-Hamadi.
\newblock A baseline for cross-database 3d human pose estimation.
\newblock {\em Sensors}, 21(11):3769, 2021.

\bibitem{tian2019contrastive}
Yonglong Tian, Dilip Krishnan, and Phillip Isola.
\newblock Contrastive multiview coding.
\newblock {\em arXiv preprint arXiv:1906.05849}, 2019.

\bibitem{tian2018rf}
Yonglong Tian, Guang-He Lee, Hao He, Chen-Yu Hsu, and Dina Katabi.
\newblock Rf-based fall monitoring using convolutional neural networks.
\newblock {\em Proceedings of the ACM on Interactive, Mobile, Wearable and
  Ubiquitous Technologies}, 2(3):1--24, 2018.

\bibitem{tian2020makes}
Yonglong Tian, Chen Sun, Ben Poole, Dilip Krishnan, Cordelia Schmid, and
  Phillip Isola.
\newblock What makes for good views for contrastive learning.
\newblock {\em arXiv preprint arXiv:2005.10243}, 2020.

\bibitem{vasisht2018duet}
Deepak Vasisht, Anubhav Jain, Chen-Yu Hsu, Zachary Kabelac, and Dina Katabi.
\newblock Duet: Estimating user position and identity in smart homes using
  intermittent and incomplete rf-data.
\newblock {\em Proceedings of the ACM on Interactive, Mobile, Wearable and
  Ubiquitous Technologies}, 2(2):1--21, 2018.

\bibitem{vincent2008extracting}
Pascal Vincent, Hugo Larochelle, Yoshua Bengio, and Pierre-Antoine Manzagol.
\newblock Extracting and composing robust features with denoising autoencoders.
\newblock In {\em Proceedings of the 25th international conference on Machine
  learning}, pages 1096--1103, 2008.

\bibitem{wang2019person}
Fei Wang, Sanping Zhou, Stanislav Panev, Jinsong Han, and Dong Huang.
\newblock Person-in-wifi: Fine-grained person perception using wifi.
\newblock In {\em Proceedings of the IEEE/CVF International Conference on
  Computer Vision}, pages 5452--5461, 2019.

\bibitem{yang2020rfid}
Chao Yang, Xuyu Wang, and Shiwen Mao.
\newblock Rfid-pose: Vision-aided three-dimensional human pose estimation with
  radio-frequency identification.
\newblock {\em IEEE Transactions on Reliability}, 2020.

\bibitem{ye2019unsupervised}
Mang Ye, Xu Zhang, Pong~C Yuen, and Shih-Fu Chang.
\newblock Unsupervised embedding learning via invariant and spreading instance
  feature.
\newblock In {\em Proceedings of the IEEE Conference on computer vision and
  pattern recognition}, pages 6210--6219, 2019.

\bibitem{yu2020human}
Jianyuan Yu, Pu Wang, Toshiaki Koike-Akino, Ye Wang, Philip~V Orlik, and
  Haijian Sun.
\newblock Human pose and seat occupancy classification with commercial mmwave
  wifi.
\newblock In {\em 2020 IEEE Globecom Workshops (GC Wkshps}, pages 1--6. IEEE,
  2020.

\bibitem{zhang2016colorful}
Richard Zhang, Phillip Isola, and Alexei~A Efros.
\newblock Colorful image colorization.
\newblock In {\em European conference on computer vision}, pages 649--666.
  Springer, 2016.

\bibitem{zhang2018latern}
Zhenyuan Zhang, Zengshan Tian, and Mu Zhou.
\newblock Latern: Dynamic continuous hand gesture recognition using fmcw radar
  sensor.
\newblock {\em IEEE Sensors Journal}, 18(8):3278--3289, 2018.

\bibitem{zhao2016emotion}
Mingmin Zhao, Fadel Adib, and Dina Katabi.
\newblock Emotion recognition using wireless signals.
\newblock In {\em Proceedings of the 22nd Annual International Conference on
  Mobile Computing and Networking}, pages 95--108, 2016.

\bibitem{zhao2018through}
Mingmin Zhao, Tianhong Li, Mohammad Abu~Alsheikh, Yonglong Tian, Hang Zhao,
  Antonio Torralba, and Dina Katabi.
\newblock Through-wall human pose estimation using radio signals.
\newblock In {\em Proceedings of the IEEE Conference on Computer Vision and
  Pattern Recognition}, pages 7356--7365, 2018.

\bibitem{zhao2019through}
Mingmin Zhao, Yingcheng Liu, Aniruddh Raghu, Tianhong Li, Hang Zhao, Antonio
  Torralba, and Dina Katabi.
\newblock Through-wall human mesh recovery using radio signals.
\newblock In {\em Proceedings of the IEEE International Conference on Computer
  Vision}, pages 10113--10122, 2019.

\bibitem{zhao2018rf}
Mingmin Zhao, Yonglong Tian, Hang Zhao, Mohammad~Abu Alsheikh, Tianhong Li,
  Rumen Hristov, Zachary Kabelac, Dina Katabi, and Antonio Torralba.
\newblock Rf-based 3d skeletons.
\newblock In {\em Proceedings of the 2018 Conference of the ACM Special
  Interest Group on Data Communication}, pages 267--281, 2018.

\bibitem{zhao2017learning}
Mingmin Zhao, Shichao Yue, Dina Katabi, Tommi~S Jaakkola, and Matt~T Bianchi.
\newblock Learning sleep stages from radio signals: A conditional adversarial
  architecture.
\newblock In {\em International Conference on Machine Learning}, pages
  4100--4109, 2017.

\bibitem{zhuang2019local}
Chengxu Zhuang, Alex~Lin Zhai, and Daniel Yamins.
\newblock Local aggregation for unsupervised learning of visual embeddings.
\newblock In {\em Proceedings of the IEEE/CVF International Conference on
  Computer Vision}, pages 6002--6012, 2019.

\end{thebibliography}
}

\end{document}